\documentclass[11pt,a4paper]{article}
\usepackage[hyperref]{emnlp2020}
\usepackage{times}
\usepackage{latexsym}

\usepackage{url}
\usepackage{times}
\usepackage{amssymb} 
\usepackage{microtype}
\usepackage{graphicx}
\usepackage{booktabs}
\usepackage{floatrow}
\usepackage{ulem}
\usepackage{booktabs}
\usepackage{subcaption} 
\usepackage{enumerate}

\aclfinalcopy 

\newcommand{\AB}[1]{#1}
\newcommand{\au}[1]{#1}

\newcommand{\ready}[1]{#1}

\newcommand{\rmbf}[1]{\textrm{\textbf{#1}}}
\newcommand{\trm}[1]{\textrm{\footnotesize{#1}}}

\DeclareUnicodeCharacter{0300}{\`o}
\DeclareUnicodeCharacter{01A1}{\^e}
\DeclareUnicodeCharacter{01B0}{\^o}

\title{UDapter: Language Adaptation for Truly Universal Dependency Parsing}

\usepackage{enumitem}

\setlist[itemize]{leftmargin=*}

\author{Ahmet \"{U}st\"{u}n \qquad Arianna Bisazza \qquad Gosse Bouma \qquad Gertjan van Noord \vspace{.2cm}
\\
University of Groningen \\
    {\tt \small \{a.ustun, a.bisazza, g.bouma, g.j.m.van.noord\}@rug.nl} 
\\}

\date{}

\begin{document}
\maketitle
\begin{abstract}
Recent advances in multilingual dependency parsing have brought the idea of a truly universal parser closer to reality.
However, cross-language interference and restrained model capacity remain major obstacles.
To address this, we propose a novel multilingual task adaptation approach based on \ready{\textit{contextual parameter generation} and \textit{adapter modules}. This approach enables to learn adapters via language embeddings while sharing model parameters across languages. It also} 
allows for an easy but effective integration of existing \textit{linguistic typology features} into the parsing network.
The resulting parser, UDapter, 
outperforms strong monolingual and multilingual baselines \ready{on the majority of} \textit{both} high-resource and low-resource (zero-shot) languages, \ready{showing the success of the proposed adaptation approach.}
Our in-depth analyses show that soft parameter sharing via typological features is key to this success.\footnote{\ready{Our code for UDapter is publicly available at \\ \url{https://github.com/ahmetustun/udapter}}}
\end{abstract}


\section{Introduction}
\label{intro}

Monolingual training of a dependency parser has been successful when relatively large treebanks are available \cite{kiperwasser2016simple,dozat2016deep}. However, for many languages, treebanks are either too small or unavailable. Therefore, multilingual models leveraging Universal Dependency annotations \cite{11234/1-2895} have drawn serious attention \cite{zhang2015hierarchical,ammar2016many,de2018parameter,kondratyuk201975}. Multilingual approaches learn generalizations across languages and share information between them, making it possible to parse a target language without supervision in that language. Moreover, multilingual models can be faster to train and easier to maintain than a large set of monolingual models.

However, scaling a multilingual model over a high number of languages can lead to sub-optimal results, especially if the training languages are typologically diverse. Often, multilingual neural models have been found to outperform their monolingual counterparts on low- and zero-resource languages due to positive transfer effects, but underperform for high-resource languages \cite{johnson2017google,arivazhagan,conneau2019unsupervised}, a problem also known as ``the curse of multilinguality''. Generally speaking, a multilingual model without language-specific supervision is likely to suffer from over-generalization and perform poorly on high-resource languages due to limited capacity compared to the monolingual baselines, as verified by our experiments on parsing. 

In this paper, we strike a good balance between maximum sharing and language-specific capacity in multilingual dependency parsing. Inspired by recently introduced parameter sharing techniques \cite{platanios2018contextual,houlsby2019parameter}, we propose a new multilingual parser, \ready{UDapter,} that learns to modify its language-specific parameters \ready{including the adapter modules,} as a function of language embeddings. This allows the model to share parameters across languages, ensuring generalization and transfer ability, but also enables language-specific parameterization in a single multilingual model. 
Furthermore, we propose not to learn language embeddings from scratch, but to leverage a mix of linguistically curated and predicted typological features as obtained from the URIEL language typology database \cite{littell2017uriel} which supports 3718 languages including all languages represented in UD. 
\ready{While the importance of typological features for cross-lingual parsing is known for both non-neural \cite{naseem2012selective,tackstrom-etal-2013-target,zhang2015hierarchical} and neural approaches \cite{ammar2016many,scholivet2019typological}, we are the first to use them \textit{effectively} as direct input to a neural parser, without manual selection, over a large number of languages in the context of zero-shot parsing where gold POS labels are not given at test time.}
In our model, typological features are crucial, leading to a substantial 
LAS increase on zero-shot languages and no loss on high-resource languages when compared to the language embeddings learned from scratch.

We train and test our model on the 13 syntactically diverse high-resource languages that were used by \newcite{kulmizev2019deep}, and also evaluate it on 30 \ready{genuinely} low-resource languages. 
\ready{Results show that UDapter significantly outperforms state-of-the-art monolingual \cite{straka-2018-udpipe} and multilingual \cite{kondratyuk201975} parsers
on \ready{most} high-resource languages and achieves overall promising improvements on zero-shot languages.}


\paragraph{Contributions}

We conduct several experiments on a large set of languages and perform thorough analyses of our model. Accordingly, we make the following contributions: 1) We apply the idea of adapter tuning \cite{rebuffi2018efficient,houlsby2019parameter} to the task of universal dependency parsing. 2) We combine adapters with the idea of contextual parameter generation \cite{platanios2018contextual}, leading to a novel language adaptation approach with state-of-the art UD parsing results. 3) We provide a simple but effective method for conditioning the language adaptation on existing typological language features, which we show is crucial for zero-shot performance.

\section{Previous Work}

%

This section presents the background of our approach.

\paragraph{Multilingual Neural Networks}

Early models in multilingual neural machine translation (NMT) designed dedicated architectures \cite{dong2015multi,firat2016multi} whilst subsequent models, from \newcite{johnson2017google} onward, added a simple language identifier to the models with the same architecture as their monolingual counterparts. More recently, multilingual NMT models have focused on maximizing transfer accuracy for low-resource language pairs, while preserving high-resource language accuracy \cite{platanios2018contextual,neubig2018rapid,aharoni2019massively,arivazhagan}, known as the (positive) transfer - (negative) interference trade-off. 
Another line of work builds massively multilingual pre-trained language models to produce contextual representation to be used in downstream tasks \cite{devlin2018bert,conneau2019unsupervised}. As the leading model, multilingual BERT (mBERT)\footnote{\url{https://github.com/google-research/bert/blob/master/multilingual.md}} \cite{devlin2018bert} which is a deep self-attention network, was trained without language-specific signals on the 104 languages with the largest  Wikipedias. It uses a shared vocabulary of 110K WordPieces \cite{wu2016google}, and has been shown to facilitate cross-lingual transfer in several applications \cite{pires2019multilingual,wu2019beto}. \ready{Concurrently to our work, \newcite{pfeiffer2020mad} have proposed to combine language and task adapters, small bottleneck layers \cite{rebuffi2018efficient,houlsby2019parameter}, to address the capacity issue which limits multilingual pre-trained models for cross-lingual transfer.}

\paragraph{Cross-Lingual Dependency Parsing}
The availability of consistent dependency treebanks in many languages \cite{mcdonald2013universal,11234/1-2895} has provided an opportunity for the study of cross-lingual parsing. Early studies trained a delexicalized parser \cite{zeman2008cross,mcdonald2013universal} on one or more source languages \ready{by using either gold or predicted POS labels \cite{tiedemann2015cross}} and applied it to target languages. Building on this, later work used additional features such as typological language properties \cite{naseem2012selective}, syntactic embeddings \cite{duong2015neural}, and  cross-lingual word clusters \cite{tackstrom2012cross}. Among lexicalized approaches, \ready{\newcite{vilares2016one} learns a bilingual parser on a corpora obtained by merging harmonized treebanks.} \newcite{ammar2016many} trains a multilingual parser using multilingual word embeddings, token-level language information, language typology features and fine-grained POS tags. More recently, based on mBERT \cite{devlin2018bert}, zero-shot transfer in dependency parsing was investigated \cite{wu2019beto,tran2019zero}.  Finally \newcite{kondratyuk201975} trained a multilingual parser on the concatenation of all available UD treebanks. 

\paragraph{\au{Language Embeddings and Typology}}
Conditioning a multilingual model on the input language is studied in NMT \cite{ha2016toward,johnson2017google}, syntactic parsing \cite{ammar2016many} and language modeling \cite{ostling2016continuous}. The goal is to embed language information in real-valued vectors in order to enrich internal representations with input language for multilingual models. 
\ready{In dependency parsing, several previous studies \cite{naseem2012selective,tackstrom-etal-2013-target,zhang2015hierarchical,ammar2016many,scholivet2019typological} have suggested that typological features are useful for the selective sharing of transfer information. 
\AB{Results, however, are mixed and often limited to a handful of manually selected features \cite{fisch-etal-2019-working,ponti2019modeling}}.
As the most similar work to ours,} \newcite{ammar2016many} uses typological features to learn language embeddings as part of training, by augmenting each input token and parsing action representation.
Unfortunately though, this technique is found to underperform the simple use of randomly initialized language embeddings (‘language IDs’).
\ready{Authors also reported that language embeddings hurt the performance of the parser in zero-shot experiments \citep[\textit{footnote~30}]{ammar2016many}.}
Our work instead demonstrates that typological features can be very effective if used with the right adaptation strategy in both supervised and zero-shot settings.
Finally, \newcite{lin2019choosing} use typological features, along with properties of the training data, to choose optimal transfer languages for various tasks, including UD parsing, in a hard manner. By contrast, we focus on a \textit{soft} parameter sharing approach to maximize generalizations within a single universal model.

\section{Proposed Model}

In this section, we present our truly universal dependency parser, UDapter.
UDapter consists of a biaffine attention layer stacked on top of the pre-trained Transformer encoder (mBERT). This is similar to \cite{wu2019beto,kondratyuk201975}, except that our mBERT layers are interleaved with special adapter layers inspired by \newcite{houlsby2019parameter}. 
While mBERT weights are frozen, biaffine attention and adapter layer weights are generated by a contextual parameter generator \cite{platanios2018contextual} that takes a language embedding as input and is updated while training on the treebanks.

Note that the proposed adaptation approach is not restricted to dependency parsing and is in principle applicable to a range of multilingual NLP tasks. We will now describe the components of our model.

\begin{figure*}[t]
\floatbox[{\capbeside\thisfloatsetup{capbesideposition={right,center},capbesidewidth=4cm}}]{figure}[\FBwidth]
{\caption{UDapter architecture with contextual parameter generator (CPG) and adapter layers. CPG takes languages embeddings projected from typological features as input and generates parameters of adapter layers and biaffine attention.}
\label{fig:model}}
{\includegraphics[width=11cm]{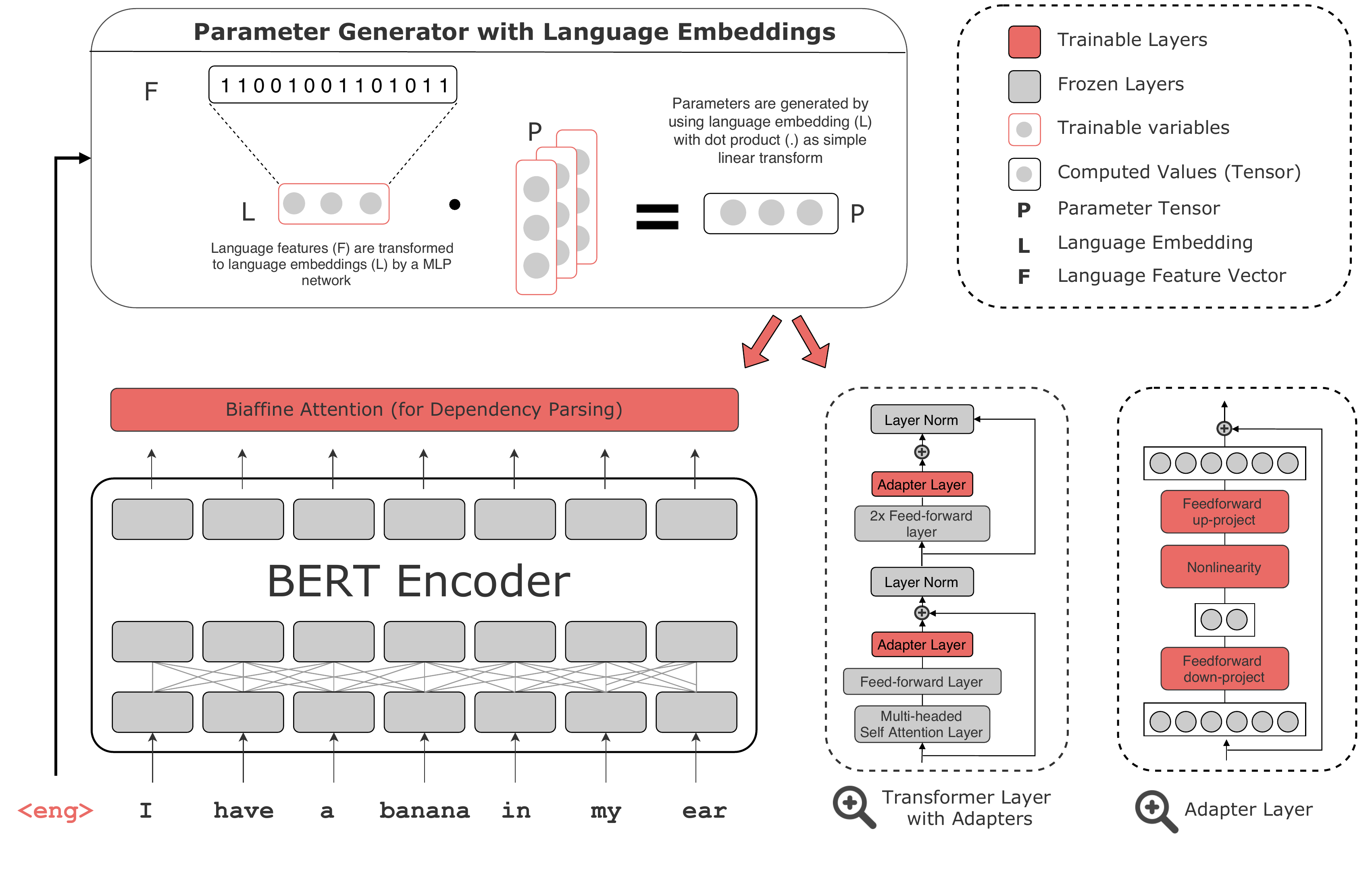}}
\end{figure*}

\subsection{Biaffine Attention Parser}
The top layer of UDapter is a graph-based biaffine attention parser proposed by \newcite{dozat2016deep}. In this model, an encoder generates an internal representation $\rmbf{r}_i$ for each word; the decoder takes $\rmbf{r}_i$ and passes it through separate feedforward layers (MLP), and finally uses deep biaffine attention to score arcs connecting a head and a tail:
\begin{eqnarray}
&\rmbf{h}^{\trm{(head)}}_i = \texttt{MLP}^{\trm{(head)}}\textrm{(}\rmbf{r}_i\textrm{)} \\
&\rmbf{h}^{\trm{(tail)}}_i = \texttt{MLP}^{\trm{(tail)}}\textrm{(}\rmbf{r}_i\textrm{)} \\
&\rmbf{s}^{\trm{(arc)}} = \texttt{Biaffine}\textrm{(}\textrm{\textbf{H}}^{\trm{(head)}},\textrm{\textbf{H}}^{\trm{(tail)}}\textrm{)}
\end{eqnarray}
\noindent Similarly,  label scores are calculated by using a biaffine classifier over two separate feedforward layers. Finally, the Chu-Liu/Edmonds algorithm \cite{chu1965shortest,edmonds1967optimum} is used to find the highest scoring valid dependency tree.

\subsection{Transformer Encoder with Adapters}

To obtain contextualized word representations, UDapter uses mBERT.
For a token $i$ in sentence $S$, BERT builds an input representation $\rmbf{w}_i$ composed by summing a WordPiece embedding $\textrm{x}_i$ \cite{wu2016google} and a position embedding $\textrm{f}_i$. Each $\rmbf{w}_i \in S$ is then passed to a stacked self-attention layers (SA) to generate the final encoder representation $\rmbf{r}_i$: 
\begin{eqnarray}
&\rmbf{w}_i = \textrm{x}_i + \textrm{f}_i \\
&\rmbf{r}_i = \texttt{SA}~\textrm{(}\rmbf{w}_i~;\Theta ^{\trm{(ad)}}\textrm{)}
\end{eqnarray}
where $\Theta ^{\trm{(ad)}}$ denotes the adapter modules. 
During training, instead of fine-tuning the whole encoder network together with the task-specific top layer, we use adapter modules \cite{rebuffi2018efficient,stickland2019bert,houlsby2019parameter}, or simply \textit{adapters},
to capture both task-specific and language-specific information.
Adapters are small modules added between layers of a pre-trained network. In adapter tuning, the weights of the original network are kept frozen, whilst the adapters are trained for a downstream task. Tuning with adapters was mainly suggested for parameter efficiency but they also act as an information module for the task or the language to be adapted \cite{pfeiffer2020mad}. In this way, the original network serves as a memory for the language(s). \ready{In UDapter, following \newcite{houlsby2019parameter}, two bottleneck adapters with two feedforward projections and a GELU nonlinearity \cite{hendrycks2016gaussian} are inserted into each transformer layer as shown in Figure \ref{fig:model}}. 
We apply adapter tuning for two reasons: 1) Each adapter module consists of only few parameters and allows to use contextual parameter generation (CPG; see \S~\ref{sec:cpg}) with a reasonable number of trainable parameters.\footnote{\ready{Due to CPG, the number of adapter parameters is multiplied by language embedding size, resulting in a larger model compared to the baseline (more details in Appendix \ref{app:imp}).}} 2) Adapters enable task-specific as well as language-specific adaptation via CPG since it keeps backbone multilingual representations as memory for all languages in pre-training, which is important for multilingual transfer. 

\begin{table*}[ht]
\small
\centering
\setlength{\tabcolsep}{3pt}
\begin{tabular}{@{}llllllllllllll@{\hskip 0.2in}cc@{}}
\toprule
              & ar    & en    & eu    & fi    & he    & hi    & it    & ja    & ko    & ru    & sv    & tr    & zh    & \textsc{hr-avg} & \textsc{lr-avg}        \\ \midrule
\multicolumn{15}{l}{\textit{Previous work:}} \\ \midrule
uuparser-bert~[1] & 81.8  & 87.6  & 79.8  & 83.9  & 85.9  & 90.8  & 91.7  & 92.1  & 84.2  & 91.0    & 86.9  & 64.9  & 83.4 & 84.9 & - \\
udpipe~[2] & 82.9 & 87.0 & 82.9 & 87.5 & 86.9 & 91.8 & 91.5 & \textbf{93.7} & 84.2 & 92.3 & 86.6 & 67.6 & 80.5 & 85.8 & - \\
udify~[3] & 82.9 & 88.5 & 81.0 & 82.1 & 88.1 & 91.5 & \textbf{93.7} & 92.1 & 74.3 & \textbf{93.1} & 89.1 & 67.4 & \textbf{83.8} & 85.2 & 34.1 \\
\midrule
\multicolumn{15}{l}{\textit{Monolingually trained (one model per language):}} \\ \midrule
mono-udify    & 83.5 & 89.4 & 81.3 & 87.3 & 87.9 & 91.1 & 93.1 & 92.5 & 84.2 & 91.9 & 88.0 & 66.0 & 82.4 & 86.0 & - \\ \midrule
\multicolumn{15}{l}{\textit{Multilingually trained (one model for all languages):}}\\ \midrule
multi-udify   & 80.1 & 88.5 & 76.4 & 85.1 & 84.4 & 89.3 & 92.0 & 90.0 & 78.0 & 89.0 & 86.2 & 62.9 & 77.8 & 83.0 & 35.3 \\
adapter-only  & 82.8 & 88.3 & 80.2 & 86.9 & 86.2 & 90.6  & 93.1 & 91.6 & 81.3 & 90.8 & 88.4 & 66.0 & 79.4 & 85.0 & 32.9 \\
udapter     & \textbf{84.4} & \textbf{89.7} & \textbf{83.3} & \textbf{89.0} & \textbf{88.8} & \textbf{92.0} & 93.5 & 92.8 & \textbf{85.9} & 92.2 & \textbf{90.3} & \textbf{69.6} & 83.2 & \textbf{87.3} & \textbf{36.5} \\ 
\bottomrule
\end{tabular}
\caption{Labelled attachment scores (LAS) on high-resource languages for baselines and UDapter. Last two columns show average LAS of 13 high-resource (\textsc{hr-avg}) and 30 low-resource (\textsc{lr-avg}) languages respectively. Previous work results are reported from \cite{kulmizev2019deep} [1] and
\cite{kondratyuk201975} [2,3].}
\label{tab:hr-langs}
\end{table*}

\begin{table*}[ht]
\small
\centering
\setlength{\tabcolsep}{2.3pt}
\begin{tabular}{@{}lcccccccccccccccccccc@{}}
\toprule
 & be  & br*   & bxr*  & cy   & fo*   & gsw*  & hsb*  & kk   & koi*  & krl*  & mdf*  & mr   & olo*  & pcm* & sa* & tl   & yo*   & yue*  & \textsc{avg} \\ \midrule
multi-udify & \textbf{80.1} & \textbf{60.5} & 26.1 & 53.6 & 68.6 & 43.6 & 53.2 & \textbf{61.9} & 20.8 & \textbf{49.2} & 24.8 & \textbf{46.4} & 42.1 & 36.1 & 19.4 & 62.7 & 41.2 & 30.5 & 45.2         \\
udapter-proxy & 69.9 & - & - & - & 64.1 & 23.7 & 44.4 & 45.1 & - & 45.6 & - & 29.6 & 41.1 & - & 15.1 & - & - & 24.5 & -         \\
udapter & 79.3 & 58.5 & \textbf{28.9} & \textbf{54.4} & \textbf{69.2} & \textbf{45.5} & \textbf{54.2} & 60.7 & \textbf{23.1} & 48.4 & \textbf{26.6} & 44.4 & \textbf{43.3} & \textbf{36.7} & \textbf{22.2} & \textbf{69.5} & \textbf{42.7} & \textbf{32.8} & \textbf{46.2}         \\ \bottomrule
\end{tabular}
\caption{Labelled attachment scores (LAS) on a subset of 30 low-resource languages. Languages with `*' are not included in mBERT training corpus. (Results for all low-resource languages, together with the chosen proxy, are given in Appendix~\ref{app:full-scores}.)}
\label{tab:lr-langs}
\end{table*}

\subsection{Contextual Parameter Generator}
\label{sec:cpg}

To control the amount of sharing across languages, we generate trainable parameters of the model using a contextual parameter generator (CPG) function inspired by \newcite{platanios2018contextual}. 
CPG enables UDapter to retain high multilingual quality without losing performance on a single language, during multi-language training. We define CPG as a function of language embeddings. Since we only train adapters and the biaffine attention 
(i.e.\ adapter tuning), the parameter generator is formalized as $\{\theta^{(ad)},~\theta^{(bf)}\} \triangleq g^{(m)}(\rmbf{l}_e)$ where $g^{(m)}$ denotes the parameter generator with language embedding $\rmbf{l}_e$, and $\theta^{(ad)}$ and $\theta^{(bf)}$ denote the parameters of adapters and biaffine attention respectively. We implement CPG as a simple linear transform of a language embedding, similar to \newcite{platanios2018contextual}, so that weights of adapters in the encoder and biaffine attention are generated by the dot product of language embeddings:
\begin{eqnarray}
g^{(m)}(\rmbf{l}_e) = (\rmbf{W}^{(\trm{ad})},\rmbf{W}^{(\trm{bf})}) \cdot \rmbf{l}_e
\end{eqnarray}

\noindent where $\rmbf{l}_e \in \mathbb{R}^\textrm{\small{M}}$, $\rmbf{~W}^{(\trm{ad})} \in \mathbb{R}^{\rmbf{\small{P}}^\trm{~(ad)}\times\textrm{\small{M}}}$, $\rmbf{~W}^{(\trm{bf})} \in \mathbb{R}^{\rmbf{\small{P}}^\trm{~(bf)}\times\textrm{\small{M}}}$, M is the language embedding size, $\rmbf{P}^\trm{~(ad)}$ and $\rmbf{P}^\trm{~(bf)}$ are the number of parameters for adapters and biaffine attention respectively.\footnote{\newcite{platanios2018contextual} also suggest to apply \textit{parameter grouping}. We have not tried that yet, but one may learn separate low-rank projections of language embeddings for the adapter parameters group and the biaffine parameters group.}
An important advantage of CPG is the easy integration of existing task or language features. 

\subsection{\AB{Typology-Based Language Embeddings}}
Soft sharing via CPG enables our model to modify its parsing decisions depending on a language embedding. While this allows UDapter to perform well on the languages in training, even if they are typologically diverse, 
information sharing is still a problem for languages not seen during training (zero-shot learning) as a language embedding is not available. Inspired by \newcite{naseem2012selective} and  \newcite{ammar2016many}, we address this problem by defining language embeddings as a function of a large set of language typological features, including syntactic and phonological features. We use a multi-layer perceptron $\texttt{MLP}^{\trm{(lang)}}$ with two feedforward layers and a \texttt{ReLU} nonlinear activation to compute a language embedding $\rmbf{l}_e$:
\begin{eqnarray}
\label{eq:mlp}
\rmbf{l}_e = \texttt{MLP}^{\trm{(lang)}}(\rmbf{l}_t)
\end{eqnarray}
\noindent where $\rmbf{l}_t$ is a typological feature vector for a language consisting of \textbf{all} 103 syntactic, 28 phonological and 158 phonetic inventory features from the URIEL language typology database \cite{littell2017uriel}. URIEL is a collection of binary features extracted from multiple typological and phylogenetic databases such as WALS (World Atlas of Language Structures) \cite{dryer2013world}, PHOIBLE \cite{moran2014phoible}, Ethnologue \cite{lewis2015ethnologue} and Glottolog \cite{glottolog}. As many feature values are not available for each language, 
we use the values predicted by \newcite{littell2017uriel} using a k-nearest neighbors approach based on average of genetic, geographical and feature distances between languages. 

\section{Experiments}
\label{sec:exp}

\paragraph{Data and Training Details}
For our training languages, we follow \newcite{kulmizev2019deep}, who selected from UD 2.3 \cite{11234/1-2895} 13 treebanks ``from different language families, with different morphological complexity, scripts, character set sizes, training sizes, domains, and with good annotation quality'' (see codes in Table~\ref{tab:hr-langs}).\footnote{
To reduce training time we cap the very large Russian Syntagrus treebank (48K sentences) to a random 15K sample.} During training, a language identifier is added to each sentence, and gold word segmentation is provided. 
We test our models on the training languages (\textit{high-resource} set), and on 30 languages that have no or very little training data (\textit{low-resource} set) in a zero-shot setup, i.e, without any training data.\footnote{For this reason, the terms `zero-shot' and `low-resource' are used interchangeably in this paper.} 
The detailed treebank list is provided in Appendix \ref{app:langs}. For evaluation, the official CoNLL 2018 Shared Task script\footnote{\url{https://universaldependencies.org/
conll18/evaluation.html}} is used to obtain LAS scores \ready{on the test set of each treebank}. 

For the encoder, we use \textit{BERT-multilingual-cased} together with its WordPiece tokenizer. Since dependency annotations are between words, we pass the BERT output corresponding to the first wordpiece per word to the biaffine parser. 
We apply the same hyper-parameter settings as \newcite{kondratyuk201975}. Additionally, we use 256 and 32 for adapter size and language embedding size respectively. In our approach, pre-trained BERT weights are frozen, and only adapters and biaffine attention are trained, thus we use the same learning rate for the whole network by applying an inverse square root learning rate decay with linear warm-up \cite{howard2018universal}. Appendix \ref{app:imp} gives the hyper-parameter details. 

\paragraph{Baselines}
We compare UDapter to the current state of the art in UD parsing:
[1]~UUparser+BERT \cite{kulmizev2019deep}, a graph-based BLSTM parser \cite{de2017raw,smith201882} using mBERT embeddings as additional features.
[2]~UDpipe \cite{straka-2018-udpipe}, a monolingually trained multi-task parser that uses pretrained word embeddings and character representations.
[3]~UDify \cite{kondratyuk201975}, the mBERT-based multi-task UD parser on which our UDapter is based, but originally trained on \textit{all} language treebanks from UD.
UDPipe scores are taken from \newcite{kondratyuk201975}.

To enable a direct comparison, we also re-train UDify on our set of 13 high-resource languages both monolingually (one treebank at a time; \textit{mono-udify}) and multilingually (on the concatenation of languages; \textit{multi-udify}).
\au{Finally, we evaluate two 
variants of our model: 1) Adapter-only has only task-specific adapter modules and no language-specific adaptation, i.e.\ no contextual parameter generator; and 
2) UDapter-proxy is trained without typology features:
 a separate language embedding is learnt from scratch for each in-training language, and for low-resource languages we use one from the same language family, if available, as proxy representation.}

Importantly, all baselines are either trained for a single language, or multilingually without any language-specific adaptation. By comparing UDapter to these parsers, we highlight its unique character that enables language specific parameterization by typological features within a multilingual framework for both supervised and zero-shot learning setup. 

\subsection{Results}
\label{sec:main-results}
Overall, UDapter outperforms the monolingual and multilingual baselines on both high-resource and zero-shot languages.
Below, we elaborate on the detailed results.

\paragraph{High-resource Languages}
Labelled Attachement Scores (LAS) on the \textit{high-resource} set are given in Table \ref{tab:hr-langs}. UDapter consistently outperforms both our monolingual and multilingual baselines in 
all languages, and beats the previous work, setting a new state of the art, in 9 out of 13 languages.
\ready{Statistical significance testing\footnote{\ready{We used \textit{paired bootstrap resampling} to check whether the difference between two models is significant (p $<$ 0.05) by using Udapi \cite{popel2017udapi}.}}
applied between UDapter and \textit{multi/mono-udify} confirms that} UDapter's performance is significantly better than the baselines in 11 out of 13 languages (all except \textit{en} and \textit{it}).

Among directly comparable baselines, \textit{multi-udify} gives the worst performance in the typologically diverse high-resource setting. This multilingual model is clearly worse than its monolingually trained counterparts \textit{mono-udify}: 83.0 \textit{vs} 86.0.
This result resounds with previous findings in multilingual NMT \cite{arivazhagan} and highlights the importance of language adaptation even when using high-quality sentence representations like those produced by mBERT. 

To understand the relevance of adapters, we also evaluate a model which has almost the same architecture as multi-udify except for the adapter modules 
and the tuning choice (frozen mBERT weights).
Interestingly, this adapter-only model considerably outperforms multi-udify (85.0 \textit{vs} 83.0), indicating that adapter modules are also effective in multilingual scenarios. 

Finally, UDapter achieves the overall best results, with consistent gains over both \textit{multi-udify} and \textit{adapter-only}, showing the importance of linguistically informed adaptation even for in-training languages. 

\begin{figure}[t]
    \includegraphics[width=\textwidth]{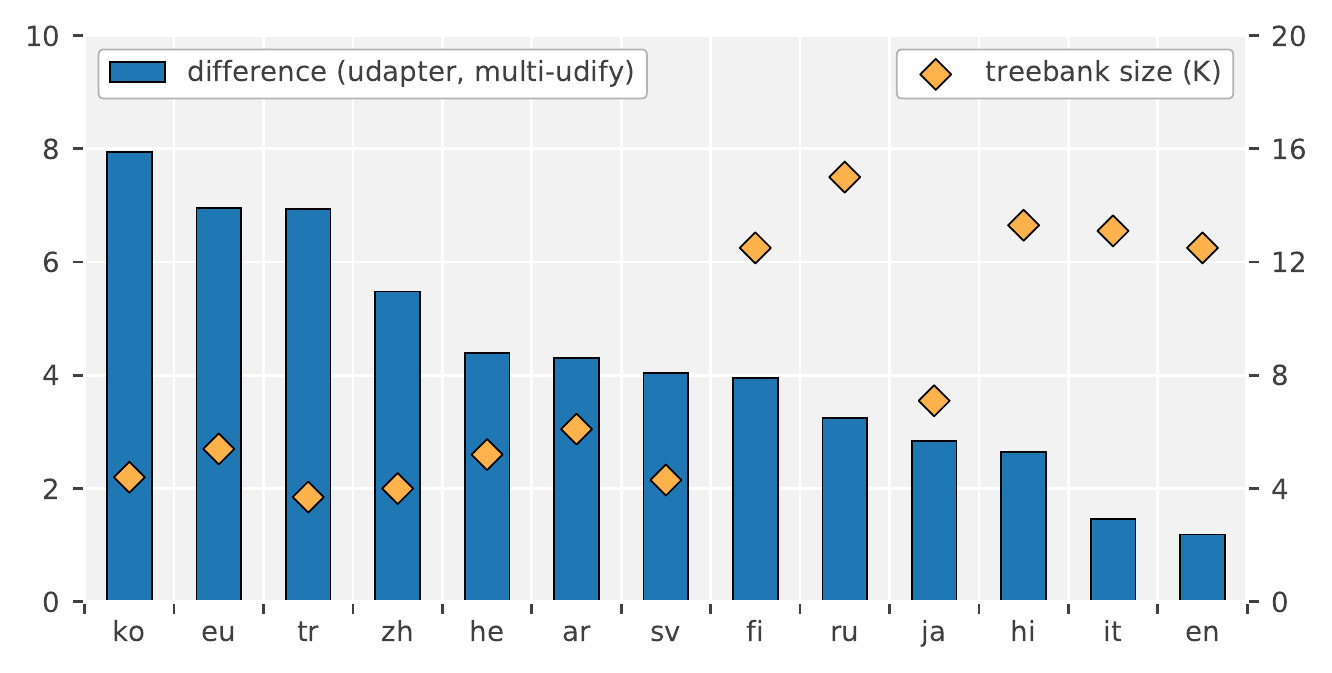}
    \caption{Difference in LAS between UDapter and multi-udify in the high-resource setting. Diamonds indicate the amount of sentences in the corresponding treebank.}
    \label{fig:hr-delta}
\end{figure}

\paragraph{Low-Resource Languages}

Average LAS on the 30 low-resource languages are shown in column \textit{lr-avg} of Table \ref{tab:hr-langs}.
\ready{Overall, UDapter slightly outperforms the multi-udify baseline (36.5 \textit{vs} 36.3)}, which shows the benefits of our approach on both in-training and zero-shot languages. 
For a closer look, Table~\ref{tab:lr-langs} provides individual results for the 18 representative languages in our low-resource set. Here we find a mixed picture: UDapter outperforms multi-udify on 13 out of 18 languages\ready{\footnote{LAS scores for all 30 languages are given in Appendix \ref{app:full-scores}. By significance testing, UDapter is \textit{significantly} better than multi-udify on 16/30 low-resource languages, which is shown in Table \ref{tab:full-lr-results}}}. Achieving improvements in the zero-shot parsing setup is very difficult, thus we believe this result is an important step towards overcoming the problem of positive/negative transfer trade-off.  

\au{Indeed, UDapter-proxy results show that choosing a \textit{proxy} language embedding from the same language family underperforms UDapter, apart from not being available for many languages. 
This indicates the importance of typological features in our approach (see \S~\ref{sec:typology} for further analysis).} 


\begin{figure}[t]
  \begin{subfigure}[b]{0.49\textwidth}
    \includegraphics[width=\textwidth]{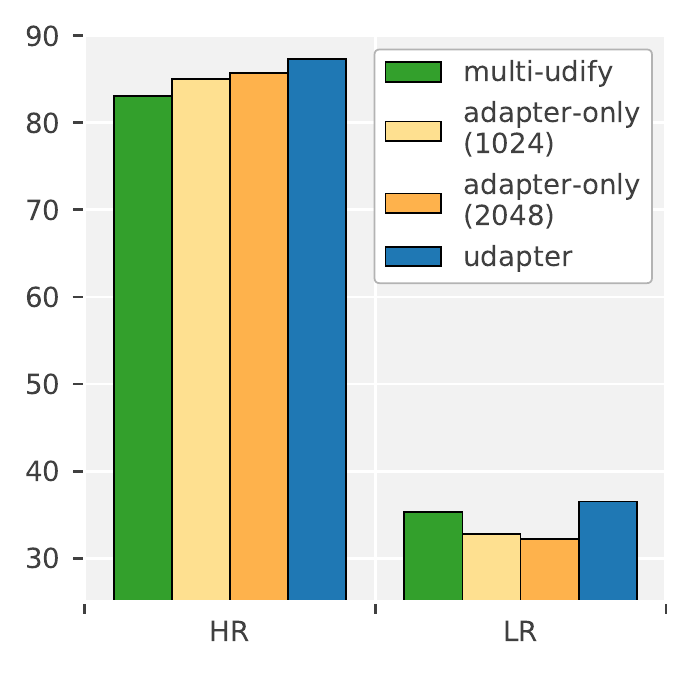}
    \caption{}
    \label{fig:adapter}
  \end{subfigure} 
  \begin{subfigure}[b]{0.49\textwidth}
    \includegraphics[width=\textwidth]{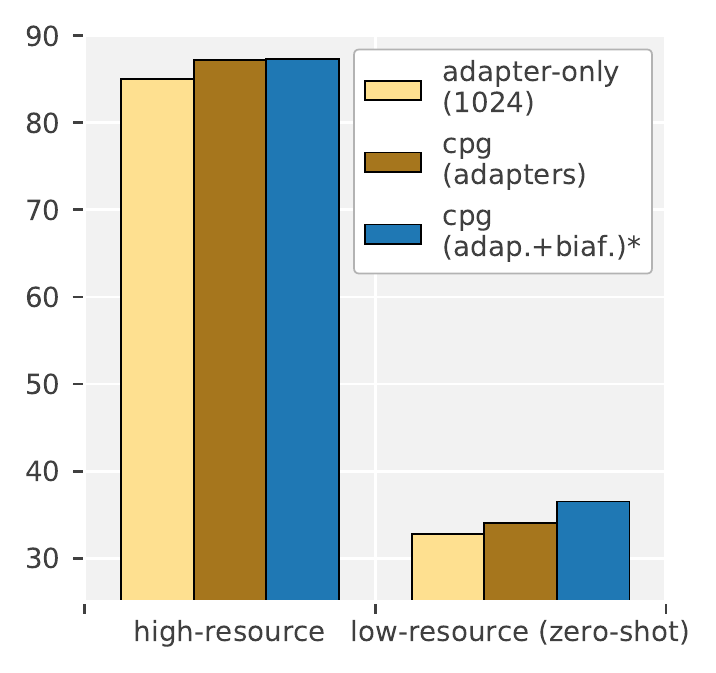}
    \caption{}
    \label{fig:ablation}
  \end{subfigure}
  \caption{Impact of different UDapter components on parsing performance (LAS): (a) adapters and adapter layer size, (b) application of contextual parameter generation to different portions of the network. In (b) the model named `cpg~(adap.+biaf.)' coincides with the full UDapter.}
  \label{fig:analysis}
\end{figure}

\section{Analysis}

In this section, we further analyse UDapter to understand its impact on different languages, and the importance of its various components.


\subsection{Which languages improve most?}
Figure \ref{fig:hr-delta} presents the LAS gain of UDapter over the multi-udify baseline for each high-resource language along with the respective treebank training size. 
To summarize, the gains are higher for languages with less training data. 
This suggests that in UDapter, useful knowledge is shared among in-training languages, which benefits low resource languages without hurting high resource ones. 

For zero-shot languages, the difference between the two models is small compared to high-resource languages (+1.2 LAS). 
While it is harder to find a trend here,
we notice that UDapter is typically beneficial for the languages \textit{not} present in the mBERT training corpus: it outperforms multi-udify in 13 out of 22 (non-mBERT) languages. 
This suggests that typological feature-based adaptation leads to improved sentence representations when the pre-trained encoder has not been exposed to a language.


\subsection{How much gain from typology?}
\label{sec:typology}
\begin{figure}[t]
 \begin{subfigure}[b]{0.49\textwidth}
    \includegraphics[width=\textwidth]{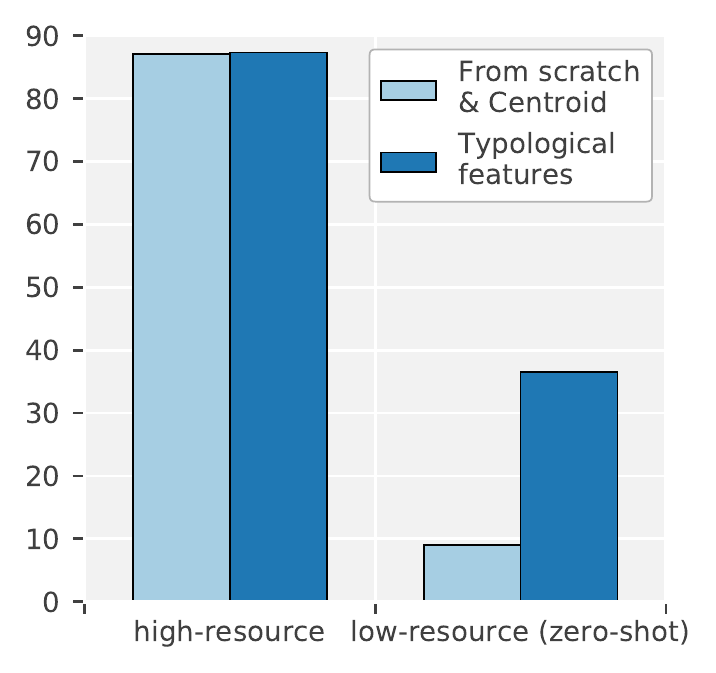}
    \caption{}
    \label{fig:typology}
  \end{subfigure} 
  \begin{subfigure}[b]{0.49\textwidth}
    \includegraphics[width=\textwidth]{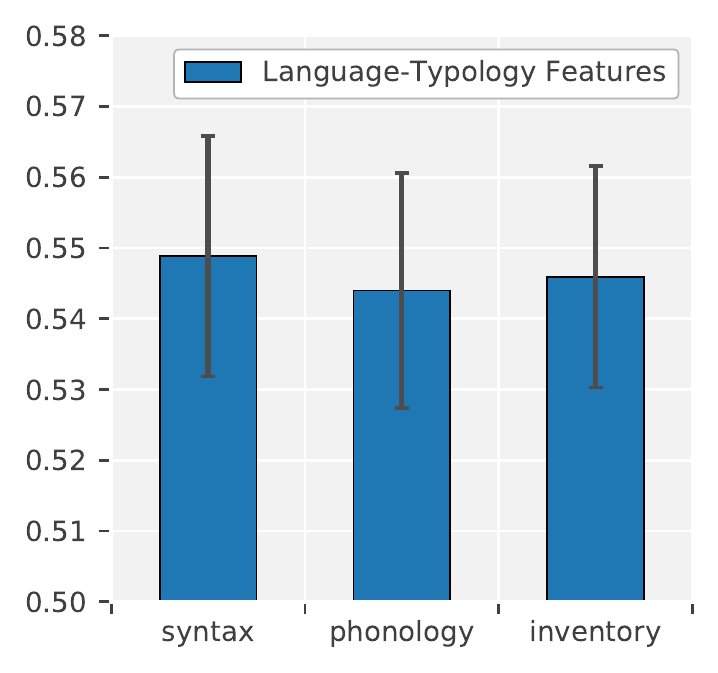}
    \caption{}
    \label{fig:feat}
  \end{subfigure}
\caption{(a) Impact of language typology features on parsing performance (LAS). (b) Average of normalized feature weights obtained from linear projection layer of the language embedding network.}
\end{figure}

UDapter learns language embeddings from syntactic, phonological and phonetic inventory features. A natural alternative to this choice is to learn language embeddings from scratch. 
For a comparison, we train a model where, for each in-training language, a separate language embedding (of the same size: 32) is initialized randomly and learned end-to-end. For the zero-shot languages 
\au{we use the average, or centroid, of all in-training language embeddings.
As shown in Figure \ref{fig:typology}, on the high-resource set, the models with and without typological features achieve very similar average LAS (87.3 and 87.1 respectively). On zero-shot languages, however, the use of centroid embedding 
performs very poorly: 9.0 \textit{vs} 36.5 average LAS score over 30 languages.}
As already discussed in \S~\ref{sec:main-results} (Table~\ref{tab:lr-langs}), using a proxy language embedding belonging to the same family as the test language, when available, also clearly underperforms UDapter.


These results confirm our expectation that a model can learn reliable language embeddings for in-training languages, however typological signals are required to obtain a robust parsing quality on zero-shot languages. 

\begin{figure*}[t!]
    \includegraphics[width=\textwidth]{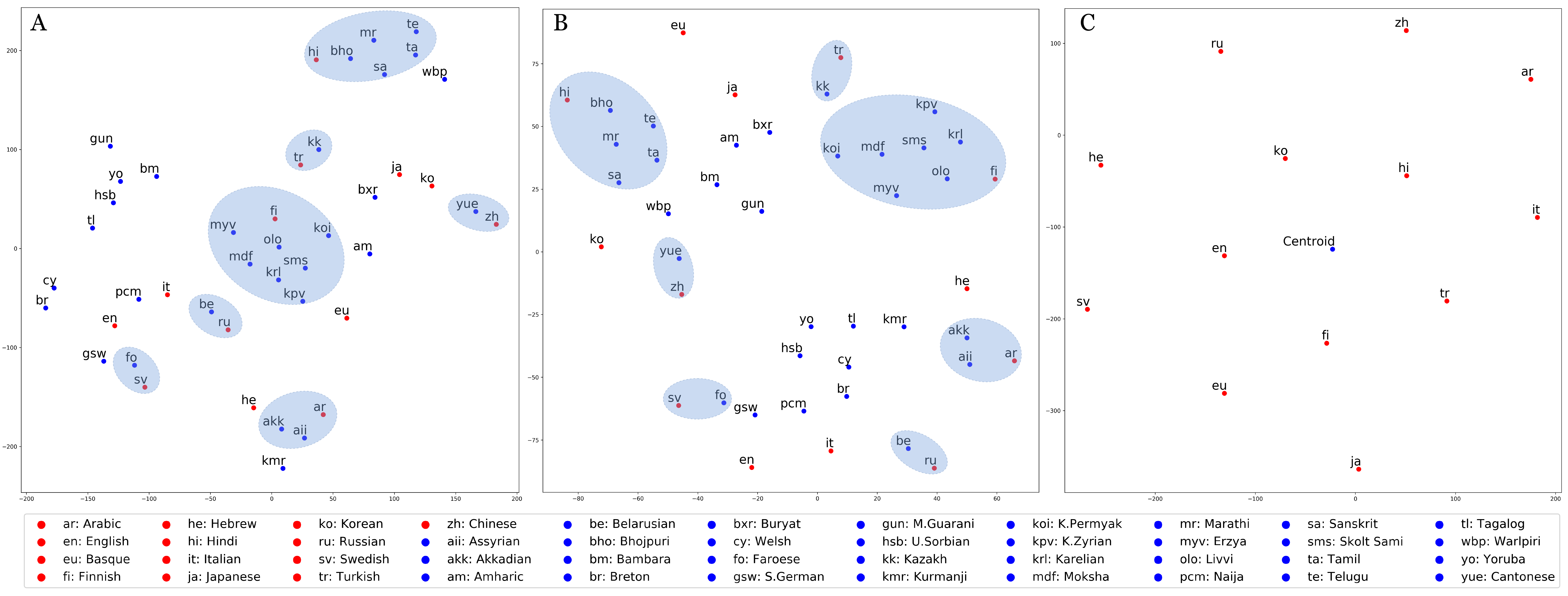}
  \caption{\au{Vector spaces for (A) language-typology feature vectors taken from URIEL, (B) language embeddings learned from typological features by UDapter, and (C) language embeddings learned without typological features.} High- and low-resource languages are indicated by red and blue dots respectively. Highlighted clusters in A and B denote sets of genetically related languages.}
  \label{fig:spaces}
\end{figure*}

\subsection{How does UDapter represent languages?}

We start by analyzing the projection weights assigned to different typological features by the first layer of the language embedding network (see eq.~\ref{eq:mlp}). Figure \ref{fig:feat} shows the averages of normalized syntactic, phonological and phonetic inventory feature \AB{weights}. Although dependency parsing is a syntactic task, the network does not only utilize syntactic features, as also observed by \newcite{lin2019choosing}, but exploits all available typological features to learn its representations. 

\ready{Next, we plot the language representations learned in UDapter by using t-SNE \cite{maaten2008visualizing}, which is similar to the analysis carried out by \newcite[figure~8]{ponti2019modeling} using the language vectors learned by \newcite{malaviya-etal-2017-learning}.}
Figure \ref{fig:spaces} illustrates 2D vector spaces generated for the typological feature vectors $\rmbf{l}_t$ (\textbf{A}) and the language embeddings $\rmbf{l}_e$ learned by UDapter with or without typological features (\textbf{B} and \textbf{C} respectively).
%
The benefits of using typological features can be understood by comparing \textbf{A} and \textbf{B}: During training, UDapter learns to project URIEL features to language embeddings in a way that is optimal for in-training language parsing quality. This leads to a different placement of the high-resource languages (red points) in the space, where many linguistic similarities are preserved (e.g. Hebrew and Arabic; European languages except Basque) but others are overruled (Japanese drifting away from Korean).
Looking at the low-resource languages (blue points) we find that typologically similar languages tend to have similar embeddings to the closest high-resource language in both \textbf{A} and \textbf{B}. In fact, most groupings of genetically related languages, such as the Indian languages (\textit{hi}-cluster) or the Uralic ones (\textit{fi}-cluster) are largely preserved across these two spaces.

Comparing \textbf{B} and \textbf{C} where language embeddings are learned from scratch, 
\AB{
the absence of typological features leads to a seemingly random space with no linguistic similarities (e.g.\ Arabic far away from Hebrew, Korean closer to English than to Japanese, etc.) and, therefore, no principled way to represent additional languages.} 

Taken together with the parsing results of \S~\ref{sec:main-results}, these plots suggest that UDapter embeddings strike a good balance between a linguistically motivated representation space and one solely optimized for in-training language accuracy.


\subsection{Is CPG really essential?}

In section~\ref{sec:main-results} we observed that adapter tuning alone (that is, without CPG) improved the multilingual baseline in the high-resource languages, but worsened it considerably in the zero-shot setup. By contrast, the addition of CPG with typological features led to the best results over all languages.
But could we have obtained similar results by simply increasing the adapter size? For instance, in multilingual MT, increasing overall model capacity of an already very large and deep architecture can be a powerful alternative to more sophisticated parameter sharing approaches \cite{arivazhagan}.
To answer this question we train another adapter-only model with doubled size (2048 instead of the 1024 used in the main experiments). 


\au{As seen in \ref{fig:adapter}, increase in model size brings a slight gain to the high-resource languages, but actually leads to a small loss in the zero-shot setup. This shows that adapters enlarge the per-language capacity for in-training languages, but at the same time they hurt generalization and zero-shot transfer. By contrast, UDapter including CPG which increases the model size by language embeddings (see Appendix \ref{app:imp} for details), outperforms both adapter-only models, confirming once more the importance of this component.}

For our last analysis (Fig.~\ref{fig:ablation}), we study soft parameter sharing via CPG on different portions of the network, namely: only on the adapter modules `cpg (adapters)' \textit{versus} on both adapters and biaffine attention `cpg (adap.+biaf.)' corresponding to the full UDapter.
Results show that most of the gain in the high-resource languages is obtained by only applying CPG on the multilingual encoder. 
On the other hand, for the low-resource languages, typological feature based parameter sharing is most important in the biaffine attention layer.
We leave further investigation of this result to future work.

\section{Conclusion}
We have presented UDapter, a multilingual dependency parsing model that learns to adapt language-specific parameters on the basis of adapter modules \cite{rebuffi2018efficient,houlsby2019parameter} and the contextual parameter generation (CPG) method \cite{platanios2018contextual} \au{which is in principle applicable to a range of multilingual NLP tasks.} While adapters provide a more general task-level adaptation, CPG 
\AB{enables language-specific adaptation, defined as a function of language embeddings projected from linguistically curated typological features.}
In this way, the model retains high per-language performance in the training data and achieves better zero-shot transfer. 

\ready{UDapter, trained on a concatenation of typologically diverse languages \cite{kulmizev2019deep}, outperforms strong monolingual \textit{and} multilingual baselines on the majority of \textit{both} high-resource and low-resource (zero-shot) languages,} 
which reflects its strong balance between per-language capacity and maximum sharing. Finally, the analyses we performed on the underlying characteristics of our model show that typological features are crucial for zero-shot languages.

\section*{Acknowledgements}
Arianna Bisazza was partly funded by the Netherlands Organization for Scientific Research (NWO) under project number 639.021.646. We would like to thank the Center for Information Technology of the University of Groningen for providing access to the Peregrine HPC cluster and the anonymous reviewers for their helpful comments.

\bibliography{emnlp2020}

\begin{thebibliography}{54}
\expandafter\ifx\csname natexlab\endcsname\relax\def\natexlab#1{#1}\fi

\bibitem[{Aharoni et~al.(2019)Aharoni, Johnson, and
  Firat}]{aharoni2019massively}
Roee Aharoni, Melvin Johnson, and Orhan Firat. 2019.
\newblock \href {https://doi.org/10.18653/v1/N19-1388} {Massively multilingual
  neural machine translation}.
\newblock In \emph{Proceedings of the 2019 Conference of the North {A}merican
  Chapter of the Association for Computational Linguistics: Human Language
  Technologies, Volume 1 (Long and Short Papers)}, pages 3874--3884.

\bibitem[{Ammar et~al.(2016)Ammar, Mulcaire, Ballesteros, Dyer, and
  Smith}]{ammar2016many}
Waleed Ammar, George Mulcaire, Miguel Ballesteros, Chris Dyer, and Noah~A.
  Smith. 2016.
\newblock \href {https://doi.org/10.1162/tacl_a_00109} {Many languages, one
  parser}.
\newblock \emph{Transactions of the Association for Computational Linguistics},
  4:431--444.

\bibitem[{Arivazhagan et~al.(2019)Arivazhagan, Bapna, Firat, Lepikhin, Johnson,
  Krikun, Chen, Cao, Foster, Cherry, Macherey, Chen, and Wu}]{arivazhagan}
Naveen Arivazhagan, Ankur Bapna, Orhan Firat, Dmitry Lepikhin, Melvin Johnson,
  Maxim Krikun, Mia~Xu Chen, Yuan Cao, George Foster, Colin Cherry, Wolfgang
  Macherey, Zhifeng Chen, and Yonghui Wu. 2019.
\newblock \href {http://arxiv.org/abs/1907.05019} {Massively multilingual
  neural machine translation in the wild: Findings and challenges}.
\newblock \emph{CoRR}, abs/1907.05019.

\bibitem[{Chu(1965)}]{chu1965shortest}
Yoeng-Jin Chu. 1965.
\newblock On the shortest arborescence of a directed graph.
\newblock \emph{Scientia Sinica}, 14:1396--1400.

\bibitem[{Conneau et~al.(2020)Conneau, Khandelwal, Goyal, Chaudhary, Wenzek,
  Guzm{\'a}n, Grave, Ott, Zettlemoyer, and Stoyanov}]{conneau2019unsupervised}
Alexis Conneau, Kartikay Khandelwal, Naman Goyal, Vishrav Chaudhary, Guillaume
  Wenzek, Francisco Guzm{\'a}n, Edouard Grave, Myle Ott, Luke Zettlemoyer, and
  Veselin Stoyanov. 2020.
\newblock \href {https://doi.org/10.18653/v1/2020.acl-main.747} {Unsupervised
  cross-lingual representation learning at scale}.
\newblock In \emph{Proceedings of the 58th Annual Meeting of the Association
  for Computational Linguistics}, pages 8440--8451.

\bibitem[{Devlin et~al.(2019)Devlin, Chang, Lee, and
  Toutanova}]{devlin2018bert}
Jacob Devlin, Ming-Wei Chang, Kenton Lee, and Kristina Toutanova. 2019.
\newblock \href {https://doi.org/10.18653/v1/N19-1423} {{BERT}: Pre-training of
  deep bidirectional transformers for language understanding}.
\newblock In \emph{Proceedings of the 2019 Conference of the North {A}merican
  Chapter of the Association for Computational Linguistics: Human Language
  Technologies, Volume 1 (Long and Short Papers)}, pages 4171--4186.

\bibitem[{Dong et~al.(2015)Dong, Wu, He, Yu, and Wang}]{dong2015multi}
Daxiang Dong, Hua Wu, Wei He, Dianhai Yu, and Haifeng Wang. 2015.
\newblock Multi-task learning for multiple language translation.
\newblock In \emph{Proceedings of the 53rd Annual Meeting of the Association
  for Computational Linguistics and the 7th International Joint Conference on
  Natural Language Processing (Volume 1: Long Papers)}, pages 1723--1732.

\bibitem[{Dozat and Manning(2017)}]{dozat2016deep}
Timothy Dozat and Christopher~D Manning. 2017.
\newblock Deep biaffine attention for neural dependency parsing.

\bibitem[{Dryer and Haspelmath(2013)}]{dryer2013world}
Matthew~S Dryer and Martin Haspelmath. 2013.
\newblock The world atlas of language structures online.

\bibitem[{Duong et~al.(2015)Duong, Cohn, Bird, and Cook}]{duong2015neural}
Long Duong, Trevor Cohn, Steven Bird, and Paul Cook. 2015.
\newblock A neural network model for low-resource universal dependency parsing.
\newblock In \emph{Proceedings of the 2015 Conference on Empirical Methods in
  Natural Language Processing}, pages 339--348.

\bibitem[{Edmonds(1967)}]{edmonds1967optimum}
Jack Edmonds. 1967.
\newblock Optimum branchings.
\newblock \emph{Journal of Research of the national Bureau of Standards B},
  71(4):233--240.

\bibitem[{Firat et~al.(2016)Firat, Cho, and Bengio}]{firat2016multi}
Orhan Firat, Kyunghyun Cho, and Yoshua Bengio. 2016.
\newblock \href {https://doi.org/10.18653/v1/N16-1101} {Multi-way, multilingual
  neural machine translation with a shared attention mechanism}.
\newblock In \emph{Proceedings of the 2016 Conference of the North {A}merican
  Chapter of the Association for Computational Linguistics: Human Language
  Technologies}, pages 866--875.

\bibitem[{Fisch et~al.(2019)Fisch, Guo, and Barzilay}]{fisch-etal-2019-working}
Adam Fisch, Jiang Guo, and Regina Barzilay. 2019.
\newblock \href {https://doi.org/10.18653/v1/D19-1574} {Working hard or hardly
  working: Challenges of integrating typology into neural dependency parsers}.
\newblock In \emph{Proceedings of the 2019 Conference on Empirical Methods in
  Natural Language Processing and the 9th International Joint Conference on
  Natural Language Processing (EMNLP-IJCNLP)}, pages 5714--5720.

\bibitem[{Ha et~al.(2016)Ha, Niehues, and Waibel}]{ha2016toward}
Thanh-Le Ha, Jan Niehues, and Alexander Waibel. 2016.
\newblock Toward multilingual neural machine translation with universal encoder
  and decoder.
\newblock \emph{arXiv preprint arXiv:1611.04798}.

\bibitem[{Hammarström et~al.(2020)Hammarström, Forkel, Haspelmath, and
  Bank}]{glottolog}
Harald Hammarström, Robert Forkel, Martin Haspelmath, and Sebastian Bank.
  2020.
\newblock \href {https://glottolog.org/ accessed 2020-04-27} {Glottolog 4.2.1}.
\newblock Max Planck Institute for the Science of Human History.

\bibitem[{Hendrycks and Gimpel(2016)}]{hendrycks2016gaussian}
Dan Hendrycks and Kevin Gimpel. 2016.
\newblock Gaussian error linear units (gelus).

\bibitem[{Houlsby et~al.(2019)Houlsby, Giurgiu, Jastrzebski, Morrone,
  De~Laroussilhe, Gesmundo, Attariyan, and Gelly}]{houlsby2019parameter}
Neil Houlsby, Andrei Giurgiu, Stanislaw Jastrzebski, Bruna Morrone, Quentin
  De~Laroussilhe, Andrea Gesmundo, Mona Attariyan, and Sylvain Gelly. 2019.
\newblock Parameter-efficient transfer learning for nlp.
\newblock In \emph{International Conference on Machine Learning}, pages
  2790--2799.

\bibitem[{Howard and Ruder(2018)}]{howard2018universal}
Jeremy Howard and Sebastian Ruder. 2018.
\newblock Universal language model fine-tuning for text classification.
\newblock In \emph{Proceedings of the 56th Annual Meeting of the Association
  for Computational Linguistics (Volume 1: Long Papers)}, pages 328--339.

\bibitem[{Johnson et~al.(2017)Johnson, Schuster, Le, Krikun, Wu, Chen, Thorat,
  Vi{\'e}gas, Wattenberg, Corrado et~al.}]{johnson2017google}
Melvin Johnson, Mike Schuster, Quoc~V Le, Maxim Krikun, Yonghui Wu, Zhifeng
  Chen, Nikhil Thorat, Fernanda Vi{\'e}gas, Martin Wattenberg, Greg Corrado,
  et~al. 2017.
\newblock Google’s multilingual neural machine translation system: Enabling
  zero-shot translation.
\newblock \emph{Transactions of the Association for Computational Linguistics},
  5:339--351.

\bibitem[{Kiperwasser and Goldberg(2016)}]{kiperwasser2016simple}
Eliyahu Kiperwasser and Yoav Goldberg. 2016.
\newblock Simple and accurate dependency parsing using bidirectional lstm
  feature representations.
\newblock \emph{Transactions of the Association for Computational Linguistics},
  4:313--327.

\bibitem[{Kondratyuk and Straka(2019)}]{kondratyuk201975}
Dan Kondratyuk and Milan Straka. 2019.
\newblock 75 languages, 1 model: Parsing universal dependencies universally.
\newblock In \emph{Proceedings of the 2019 Conference on Empirical Methods in
  Natural Language Processing and the 9th International Joint Conference on
  Natural Language Processing (EMNLP-IJCNLP)}, pages 2779--2795.

\bibitem[{Kulmizev et~al.(2019)Kulmizev, de~Lhoneux, Gontrum, Fano, and
  Nivre}]{kulmizev2019deep}
Artur Kulmizev, Miryam de~Lhoneux, Johannes Gontrum, Elena Fano, and Joakim
  Nivre. 2019.
\newblock \href {https://doi.org/10.18653/v1/D19-1277} {Deep contextualized
  word embeddings in transition-based and graph-based dependency parsing - a
  tale of two parsers revisited}.
\newblock In \emph{Proceedings of the 2019 Conference on Empirical Methods in
  Natural Language Processing and the 9th International Joint Conference on
  Natural Language Processing (EMNLP-IJCNLP)}, pages 2755--2768.

\bibitem[{Lewis et~al.(2015)Lewis, Simons, and Fennig}]{lewis2015ethnologue}
M~Paul Lewis, Gary~F Simons, and CD~Fennig. 2015.
\newblock Ethnologue: Languages of the world [eighteenth.
\newblock \emph{Dallas, Texas: SIL International}.

\bibitem[{de~Lhoneux et~al.(2018)de~Lhoneux, Bjerva, Augenstein, and
  S{\o}gaard}]{de2018parameter}
Miryam de~Lhoneux, Johannes Bjerva, Isabelle Augenstein, and Anders S{\o}gaard.
  2018.
\newblock Parameter sharing between dependency parsers for related languages.
\newblock In \emph{Proceedings of the 2018 Conference on Empirical Methods in
  Natural Language Processing}, pages 4992--4997.

\bibitem[{de~Lhoneux et~al.(2017)de~Lhoneux, Shao, Basirat, Kiperwasser,
  Stymne, Goldberg, and Nivre}]{de2017raw}
Miryam de~Lhoneux, Yan Shao, Ali Basirat, Eliyahu Kiperwasser, Sara Stymne,
  Yoav Goldberg, and Joakim Nivre. 2017.
\newblock From raw text to universal dependencies-look, no tags!
\newblock In \emph{Proceedings of the CoNLL 2017 Shared Task: Multilingual
  Parsing from Raw Text to Universal Dependencies}, pages 207--217.

\bibitem[{Lin et~al.(2019)Lin, Chen, Lee, Li, Zhang, Xia, Rijhwani, He, Zhang,
  Ma et~al.}]{lin2019choosing}
Yu-Hsiang Lin, Chian-Yu Chen, Jean Lee, Zirui Li, Yuyan Zhang, Mengzhou Xia,
  Shruti Rijhwani, Junxian He, Zhisong Zhang, Xuezhe Ma, et~al. 2019.
\newblock Choosing transfer languages for cross-lingual learning.
\newblock In \emph{Proceedings of the 57th Annual Meeting of the Association
  for Computational Linguistics}, pages 3125--3135.

\bibitem[{Littell et~al.(2017)Littell, Mortensen, Lin, Kairis, Turner, and
  Levin}]{littell2017uriel}
Patrick Littell, David~R Mortensen, Ke~Lin, Katherine Kairis, Carlisle Turner,
  and Lori Levin. 2017.
\newblock Uriel and lang2vec: Representing languages as typological,
  geographical, and phylogenetic vectors.
\newblock In \emph{Proceedings of the 15th Conference of the European Chapter
  of the Association for Computational Linguistics: Volume 2, Short Papers},
  pages 8--14.

\bibitem[{Maaten and Hinton(2008)}]{maaten2008visualizing}
Laurens van~der Maaten and Geoffrey Hinton. 2008.
\newblock Visualizing data using t-sne.
\newblock \emph{Journal of machine learning research}, 9(Nov):2579--2605.

\bibitem[{Malaviya et~al.(2017)Malaviya, Neubig, and
  Littell}]{malaviya-etal-2017-learning}
Chaitanya Malaviya, Graham Neubig, and Patrick Littell. 2017.
\newblock \href {https://doi.org/10.18653/v1/D17-1268} {Learning language
  representations for typology prediction}.
\newblock In \emph{Proceedings of the 2017 Conference on Empirical Methods in
  Natural Language Processing}, pages 2529--2535.

\bibitem[{McDonald et~al.(2013)McDonald, Nivre, Quirmbach-Brundage, Goldberg,
  Das, Ganchev, Hall, Petrov, Zhang, T{\"a}ckstr{\"o}m
  et~al.}]{mcdonald2013universal}
Ryan McDonald, Joakim Nivre, Yvonne Quirmbach-Brundage, Yoav Goldberg, Dipanjan
  Das, Kuzman Ganchev, Keith Hall, Slav Petrov, Hao Zhang, Oscar
  T{\"a}ckstr{\"o}m, et~al. 2013.
\newblock Universal dependency annotation for multilingual parsing.
\newblock In \emph{Proceedings of the 51st Annual Meeting of the Association
  for Computational Linguistics (Volume 2: Short Papers)}, pages 92--97.

\bibitem[{Moran et~al.(2014)Moran, McCloy, and Wright}]{moran2014phoible}
Steven Moran, Daniel McCloy, and Richard Wright. 2014.
\newblock Phoible online.

\bibitem[{Naseem et~al.(2012)Naseem, Barzilay, and
  Globerson}]{naseem2012selective}
Tahira Naseem, Regina Barzilay, and Amir Globerson. 2012.
\newblock Selective sharing for multilingual dependency parsing.
\newblock In \emph{Proceedings of the 50th Annual Meeting of the Association
  for Computational Linguistics: Long Papers-Volume 1}, pages 629--637.
  Association for Computational Linguistics.

\bibitem[{Neubig and Hu(2018)}]{neubig2018rapid}
Graham Neubig and Junjie Hu. 2018.
\newblock Rapid adaptation of neural machine translation to new languages.
\newblock In \emph{Proceedings of the 2018 Conference on Empirical Methods in
  Natural Language Processing}, pages 875--880.

\bibitem[{Nivre et~al.(2018)Nivre, Abrams, Agi{\'c}, Ahrenberg, Antonsen,
  Aplonova, Aranzabe, Arutie, Asahara, Ateyah, Attia, Atutxa, Augustinus,
  Badmaeva, Ballesteros, Banerjee, Bank, Barbu~Mititelu, Basmov, Bauer,
  Bellato, Bengoetxea, Berzak, Bhat, Bhat, Biagetti, Bick, Blokland, Bobicev,
  B{\"o}rstell, Bosco, Bouma, Bowman, Boyd, Burchardt, Candito, Caron, Caron,
  Cebiro{\u g}lu~Eryi{\u g}it, Cecchini, Celano, {\v C}{\'e}pl{\"o}, Cetin,
  Chalub, Choi, Cho, Chun, Cinkov{\'a}, Collomb, {\c C}{\"o}ltekin, Connor,
  Courtin, Davidson, de~Marneffe, de~Paiva, Diaz~de Ilarraza, Dickerson, Dirix,
  Dobrovoljc, Dozat, Droganova, Dwivedi, Eli, Elkahky, Ephrem, Erjavec,
  Etienne, Farkas, Fernandez~Alcalde, Foster, Freitas, Gajdo{\v s}ov{\'a},
  Galbraith, Garcia, G{\"a}rdenfors, Garza, Gerdes, Ginter, Goenaga, Gojenola,
  G{\"o}k{\i}rmak, Goldberg, G{\'o}mez~Guinovart, Gonz{\'a}les~Saavedra,
  Grioni, Gr{\=u}z{\={\i}}tis, Guillaume, Guillot-Barbance, Habash, Haji{\v c},
  Haji{\v c}~jr., H{\`a}~M{\~y}, Han, Harris, Haug, Hladk{\'a}, Hlav{\'a}{\v
  c}ov{\'a}, Hociung, Hohle, Hwang, Ion, Irimia, Ishola, Jel{\'{\i}}nek,
  Johannsen, J{\o}rgensen, Ka{\c s}{\i}kara, Kahane, Kanayama, Kanerva, Katz,
  Kayadelen, Kenney, Kettnerov{\'a}, Kirchner, Kopacewicz, Kotsyba, Krek, Kwak,
  Laippala, Lambertino, Lam, Lando, Larasati, Lavrentiev, Lee,
  L{\^e}~H{\`{\^o}}ng, Lenci, Lertpradit, Leung, Li, Li, Li, Lim, Ljube{\v
  s}i{\'c}, Loginova, Lyashevskaya, Lynn, Macketanz, Makazhanov, Mandl,
  Manning, Manurung, M{\u a}r{\u a}nduc, Mare{\v c}ek, Marheinecke,
  Mart{\'{\i}}nez~Alonso, Martins, Ma{\v s}ek, Matsumoto, {McDonald}, Mendon{\c
  c}a, Miekka, Misirpashayeva, Missil{\"a}, Mititelu, Miyao, Montemagni, More,
  Moreno~Romero, Mori, Mori, Mortensen, Moskalevskyi, Muischnek, Murawaki,
  M{\"u}{\"u}risep, Nainwani, Navarro~Hor{\~n}iacek, Nedoluzhko, Ne{\v
  s}pore-B{\=e}rzkalne, Nguy{\~{\^e}}n~Th{\d i}, Nguy{\~{\^e}}n Th{\d i}~Minh,
  Nikolaev, Nitisaroj, Nurmi, Ojala, Ol{\'u}{\`o}kun, Omura, Osenova,
  {\"O}stling, {\O}vrelid, Partanen, Pascual, Passarotti, Patejuk,
  Paulino-Passos, Peng, Perez, Perrier, Petrov, Piitulainen, Pitler, Plank,
  Poibeau, Popel, Pretkalni{\c n}a, Pr{\'e}vost, Prokopidis,
  Przepi{\'o}rkowski, Puolakainen, Pyysalo, R{\"a}{\"a}bis, Rademaker,
  Ramasamy, Rama, Ramisch, Ravishankar, Real, Reddy, Rehm, Rie{\ss}ler,
  Rinaldi, Rituma, Rocha, Romanenko, Rosa, Rovati, Roșca, Rudina, Rueter,
  Sadde, Sagot, Saleh, Samard{\v z}i{\'c}, Samson, Sanguinetti, Saul{\={\i}}te,
  Sawanakunanon, Schneider, Schuster, Seddah, Seeker, Seraji, Shen, Shimada,
  Shohibussirri, Sichinava, Silveira, Simi, Simionescu, Simk{\'o}, {\v
  S}imkov{\'a}, Simov, Smith, Soares-Bastos, Spadine, Stella, Straka,
  Strnadov{\'a}, Suhr, Sulubacak, Sz{\'a}nt{\'o}, Taji, Takahashi, Tanaka,
  Tellier, Trosterud, Trukhina, Tsarfaty, Tyers, Uematsu, Ure{\v s}ov{\'a},
  Uria, Uszkoreit, Vajjala, van Niekerk, van Noord, Varga, Villemonte de~la
  Clergerie, Vincze, Wallin, Wang, Washington, Williams, Wir{\'e}n,
  Woldemariam, Wong, Yan, Yavrumyan, Yu, {\v Z}abokrtsk{\'y}, Zeldes, Zeman,
  Zhang, and Zhu}]{11234/1-2895}
Joakim Nivre, Mitchell Abrams, {\v Z}eljko Agi{\'c}, Lars Ahrenberg, Lene
  Antonsen, Katya Aplonova, Maria~Jesus Aranzabe, Gashaw Arutie, Masayuki
  Asahara, Luma Ateyah, Mohammed Attia, Aitziber Atutxa, Liesbeth Augustinus,
  Elena Badmaeva, Miguel Ballesteros, Esha Banerjee, Sebastian Bank, Verginica
  Barbu~Mititelu, Victoria Basmov, John Bauer, Sandra Bellato, Kepa Bengoetxea,
  Yevgeni Berzak, Irshad~Ahmad Bhat, Riyaz~Ahmad Bhat, Erica Biagetti, Eckhard
  Bick, Rogier Blokland, Victoria Bobicev, Carl B{\"o}rstell, Cristina Bosco,
  Gosse Bouma, Sam Bowman, Adriane Boyd, Aljoscha Burchardt, Marie Candito,
  Bernard Caron, Gauthier Caron, G{\"u}l{\c s}en Cebiro{\u g}lu~Eryi{\u g}it,
  Flavio~Massimiliano Cecchini, Giuseppe G.~A. Celano, Slavom{\'{\i}}r {\v
  C}{\'e}pl{\"o}, Savas Cetin, Fabricio Chalub, Jinho Choi, Yongseok Cho,
  Jayeol Chun, Silvie Cinkov{\'a}, Aur{\'e}lie Collomb, {\c C}a{\u g}r{\i} {\c
  C}{\"o}ltekin, Miriam Connor, Marine Courtin, Elizabeth Davidson,
  Marie-Catherine de~Marneffe, Valeria de~Paiva, Arantza Diaz~de Ilarraza,
  Carly Dickerson, Peter Dirix, Kaja Dobrovoljc, Timothy Dozat, Kira Droganova,
  Puneet Dwivedi, Marhaba Eli, Ali Elkahky, Binyam Ephrem, Toma{\v z} Erjavec,
  Aline Etienne, Rich{\'a}rd Farkas, Hector Fernandez~Alcalde, Jennifer Foster,
  Cl{\'a}udia Freitas, Katar{\'{\i}}na Gajdo{\v s}ov{\'a}, Daniel Galbraith,
  Marcos Garcia, Moa G{\"a}rdenfors, Sebastian Garza, Kim Gerdes, Filip Ginter,
  Iakes Goenaga, Koldo Gojenola, Memduh G{\"o}k{\i}rmak, Yoav Goldberg, Xavier
  G{\'o}mez~Guinovart, Berta Gonz{\'a}les~Saavedra, Matias Grioni, Normunds
  Gr{\=u}z{\={\i}}tis, Bruno Guillaume, C{\'e}line Guillot-Barbance, Nizar
  Habash, Jan Haji{\v c}, Jan Haji{\v c}~jr., Linh H{\`a}~M{\~y}, Na-Rae Han,
  Kim Harris, Dag Haug, Barbora Hladk{\'a}, Jaroslava Hlav{\'a}{\v c}ov{\'a},
  Florinel Hociung, Petter Hohle, Jena Hwang, Radu Ion, Elena Irimia, {\d
  O}l{\'a}j{\'{\i}}d{\'e} Ishola, Tom{\'a}{\v s} Jel{\'{\i}}nek, Anders
  Johannsen, Fredrik J{\o}rgensen, H{\"u}ner Ka{\c s}{\i}kara, Sylvain Kahane,
  Hiroshi Kanayama, Jenna Kanerva, Boris Katz, Tolga Kayadelen, Jessica Kenney,
  V{\'a}clava Kettnerov{\'a}, Jesse Kirchner, Kamil Kopacewicz, Natalia
  Kotsyba, Simon Krek, Sookyoung Kwak, Veronika Laippala, Lorenzo Lambertino,
  Lucia Lam, Tatiana Lando, Septina~Dian Larasati, Alexei Lavrentiev, John Lee,
  Phương L{\^e}~H{\`{\^o}}ng, Alessandro Lenci, Saran Lertpradit, Herman
  Leung, Cheuk~Ying Li, Josie Li, Keying Li, {KyungTae} Lim, Nikola Ljube{\v
  s}i{\'c}, Olga Loginova, Olga Lyashevskaya, Teresa Lynn, Vivien Macketanz,
  Aibek Makazhanov, Michael Mandl, Christopher Manning, Ruli Manurung, C{\u
  a}t{\u a}lina M{\u a}r{\u a}nduc, David Mare{\v c}ek, Katrin Marheinecke,
  H{\'e}ctor Mart{\'{\i}}nez~Alonso, Andr{\'e} Martins, Jan Ma{\v s}ek, Yuji
  Matsumoto, Ryan {McDonald}, Gustavo Mendon{\c c}a, Niko Miekka, Margarita
  Misirpashayeva, Anna Missil{\"a}, C{\u a}t{\u a}lin Mititelu, Yusuke Miyao,
  Simonetta Montemagni, Amir More, Laura Moreno~Romero, Keiko~Sophie Mori,
  Shinsuke Mori, Bjartur Mortensen, Bohdan Moskalevskyi, Kadri Muischnek, Yugo
  Murawaki, Kaili M{\"u}{\"u}risep, Pinkey Nainwani, Juan~Ignacio
  Navarro~Hor{\~n}iacek, Anna Nedoluzhko, Gunta Ne{\v s}pore-B{\=e}rzkalne,
  Lương Nguy{\~{\^e}}n~Th{\d i}, Huy{\`{\^e}}n Nguy{\~{\^e}}n Th{\d i}~Minh,
  Vitaly Nikolaev, Rattima Nitisaroj, Hanna Nurmi, Stina Ojala, Ad{\'e}day{\d
  o}̀ Ol{\'u}{\`o}kun, Mai Omura, Petya Osenova, Robert {\"O}stling, Lilja
  {\O}vrelid, Niko Partanen, Elena Pascual, Marco Passarotti, Agnieszka
  Patejuk, Guilherme Paulino-Passos, Siyao Peng, Cenel-Augusto Perez, Guy
  Perrier, Slav Petrov, Jussi Piitulainen, Emily Pitler, Barbara Plank, Thierry
  Poibeau, Martin Popel, Lauma Pretkalni{\c n}a, Sophie Pr{\'e}vost, Prokopis
  Prokopidis, Adam Przepi{\'o}rkowski, Tiina Puolakainen, Sampo Pyysalo,
  Andriela R{\"a}{\"a}bis, Alexandre Rademaker, Loganathan Ramasamy, Taraka
  Rama, Carlos Ramisch, Vinit Ravishankar, Livy Real, Siva Reddy, Georg Rehm,
  Michael Rie{\ss}ler, Larissa Rinaldi, Laura Rituma, Luisa Rocha, Mykhailo
  Romanenko, Rudolf Rosa, Davide Rovati, Valentin Roșca, Olga Rudina, Jack
  Rueter, Shoval Sadde, Beno{\^{\i}}t Sagot, Shadi Saleh, Tanja Samard{\v
  z}i{\'c}, Stephanie Samson, Manuela Sanguinetti, Baiba Saul{\={\i}}te, Yanin
  Sawanakunanon, Nathan Schneider, Sebastian Schuster, Djam{\'e} Seddah,
  Wolfgang Seeker, Mojgan Seraji, Mo~Shen, Atsuko Shimada, Muh Shohibussirri,
  Dmitry Sichinava, Natalia Silveira, Maria Simi, Radu Simionescu, Katalin
  Simk{\'o}, M{\'a}ria {\v S}imkov{\'a}, Kiril Simov, Aaron Smith, Isabela
  Soares-Bastos, Carolyn Spadine, Antonio Stella, Milan Straka, Jana
  Strnadov{\'a}, Alane Suhr, Umut Sulubacak, Zsolt Sz{\'a}nt{\'o}, Dima Taji,
  Yuta Takahashi, Takaaki Tanaka, Isabelle Tellier, Trond Trosterud, Anna
  Trukhina, Reut Tsarfaty, Francis Tyers, Sumire Uematsu, Zde{\v n}ka Ure{\v
  s}ov{\'a}, Larraitz Uria, Hans Uszkoreit, Sowmya Vajjala, Daniel van Niekerk,
  Gertjan van Noord, Viktor Varga, Eric Villemonte de~la Clergerie, Veronika
  Vincze, Lars Wallin, Jing~Xian Wang, Jonathan~North Washington, Seyi
  Williams, Mats Wir{\'e}n, Tsegay Woldemariam, Tak-sum Wong, Chunxiao Yan,
  Marat~M. Yavrumyan, Zhuoran Yu, Zden{\v e}k {\v Z}abokrtsk{\'y}, Amir Zeldes,
  Daniel Zeman, Manying Zhang, and Hanzhi Zhu. 2018.
\newblock \href {http://hdl.handle.net/11234/1-2895} {Universal dependencies
  2.3}.
\newblock {LINDAT}/{CLARIAH}-{CZ} digital library at the Institute of Formal
  and Applied Linguistics ({{\'U}FAL}), Faculty of Mathematics and Physics,
  Charles University.

\bibitem[{{\"O}stling and Tiedemann(2017)}]{ostling2016continuous}
Robert {\"O}stling and J{\"o}rg Tiedemann. 2017.
\newblock \href {https://www.aclweb.org/anthology/E17-2102} {Continuous
  multilinguality with language vectors}.
\newblock In \emph{Proceedings of the 15th Conference of the {E}uropean Chapter
  of the Association for Computational Linguistics: Volume 2, Short Papers},
  pages 644--649.

\bibitem[{Pfeiffer et~al.(2020)Pfeiffer, Vuli{\'c}, Gurevych, and
  Ruder}]{pfeiffer2020mad}
Jonas Pfeiffer, Ivan Vuli{\'c}, Iryna Gurevych, and Sebastian Ruder. 2020.
\newblock Mad-x: An adapter-based framework for multi-task cross-lingual
  transfer.
\newblock In \emph{Proceedings of the 2020 Conference on Empirical Methods in
  Natural Language Processing}.

\bibitem[{Pires et~al.(2019)Pires, Schlinger, and
  Garrette}]{pires2019multilingual}
Telmo Pires, Eva Schlinger, and Dan Garrette. 2019.
\newblock How multilingual is multilingual bert?
\newblock In \emph{Proceedings of the 57th Annual Meeting of the Association
  for Computational Linguistics}, pages 4996--5001.

\bibitem[{Platanios et~al.(2018)Platanios, Sachan, Neubig, and
  Mitchell}]{platanios2018contextual}
Emmanouil~Antonios Platanios, Mrinmaya Sachan, Graham Neubig, and Tom Mitchell.
  2018.
\newblock Contextual parameter generation for universal neural machine
  translation.
\newblock In \emph{Proceedings of the 2018 Conference on Empirical Methods in
  Natural Language Processing}, pages 425--435.

\bibitem[{Ponti et~al.(2019)Ponti, O’horan, Berzak, Vuli{\'c}, Reichart,
  Poibeau, Shutova, and Korhonen}]{ponti2019modeling}
Edoardo~Maria Ponti, Helen O’horan, Yevgeni Berzak, Ivan Vuli{\'c}, Roi
  Reichart, Thierry Poibeau, Ekaterina Shutova, and Anna Korhonen. 2019.
\newblock Modeling language variation and universals: A survey on typological
  linguistics for natural language processing.
\newblock \emph{Computational Linguistics}, 45(3):559--601.

\bibitem[{Popel et~al.(2017)Popel, {\v{Z}}abokrtsk{\`y}, and
  Vojtek}]{popel2017udapi}
Martin Popel, Zden{\v{e}}k {\v{Z}}abokrtsk{\`y}, and Martin Vojtek. 2017.
\newblock Udapi: Universal api for universal dependencies.
\newblock In \emph{Proceedings of the NoDaLiDa 2017 Workshop on Universal
  Dependencies (UDW 2017)}, pages 96--101.

\bibitem[{Rebuffi et~al.(2018)Rebuffi, Bilen, and
  Vedaldi}]{rebuffi2018efficient}
Sylvestre-Alvise Rebuffi, Hakan Bilen, and Andrea Vedaldi. 2018.
\newblock Efficient parametrization of multi-domain deep neural networks.
\newblock In \emph{Proceedings of the IEEE Conference on Computer Vision and
  Pattern Recognition}, pages 8119--8127.

\bibitem[{Scholivet et~al.(2019)Scholivet, Dary, Nasr, Favre, and
  Ramisch}]{scholivet2019typological}
Manon Scholivet, Franck Dary, Alexis Nasr, Benoit Favre, and Carlos Ramisch.
  2019.
\newblock Typological features for multilingual delexicalised dependency
  parsing.
\newblock In \emph{Proceedings of the 2019 Conference of the North American
  Chapter of the Association for Computational Linguistics: Human Language
  Technologies, Volume 1 (Long and Short Papers)}, pages 3919--3930.

\bibitem[{Smith et~al.(2018)Smith, Bohnet, de~Lhoneux, Nivre, Shao, and
  Stymne}]{smith201882}
Aaron Smith, Bernd Bohnet, Miryam de~Lhoneux, Joakim Nivre, Yan Shao, and Sara
  Stymne. 2018.
\newblock 82 treebanks, 34 models: Universal dependency parsing with
  multi-treebank models.
\newblock \emph{CoNLL 2018}, page 113.

\bibitem[{Stickland and Murray(2019)}]{stickland2019bert}
Asa~Cooper Stickland and Iain Murray. 2019.
\newblock Bert and pals: Projected attention layers for efficient adaptation in
  multi-task learning.
\newblock In \emph{International Conference on Machine Learning}, pages
  5986--5995.

\bibitem[{Straka(2018)}]{straka-2018-udpipe}
Milan Straka. 2018.
\newblock \href {https://doi.org/10.18653/v1/K18-2020} {{UDP}ipe 2.0 prototype
  at {C}o{NLL} 2018 {UD} shared task}.
\newblock In \emph{Proceedings of the {C}o{NLL} 2018 Shared Task: Multilingual
  Parsing from Raw Text to Universal Dependencies}, pages 197--207.

\bibitem[{T{\"a}ckstr{\"o}m et~al.(2013)T{\"a}ckstr{\"o}m, McDonald, and
  Nivre}]{tackstrom-etal-2013-target}
Oscar T{\"a}ckstr{\"o}m, Ryan McDonald, and Joakim Nivre. 2013.
\newblock \href {https://www.aclweb.org/anthology/N13-1126} {Target language
  adaptation of discriminative transfer parsers}.
\newblock In \emph{Proceedings of the 2013 Conference of the North {A}merican
  Chapter of the Association for Computational Linguistics: Human Language
  Technologies}, pages 1061--1071.

\bibitem[{T{\"a}ckstr{\"o}m et~al.(2012)T{\"a}ckstr{\"o}m, McDonald, and
  Uszkoreit}]{tackstrom2012cross}
Oscar T{\"a}ckstr{\"o}m, Ryan McDonald, and Jakob Uszkoreit. 2012.
\newblock Cross-lingual word clusters for direct transfer of linguistic
  structure.
\newblock In \emph{Proceedings of the 2012 conference of the North American
  chapter of the association for computational linguistics: Human language
  technologies}, pages 477--487. Association for Computational Linguistics.

\bibitem[{Tiedemann(2015)}]{tiedemann2015cross}
J{\"o}rg Tiedemann. 2015.
\newblock Cross-lingual dependency parsing with universal dependencies and
  predicted pos labels.
\newblock \emph{Depling 2015}, page 340.

\bibitem[{Tran and Bisazza(2019)}]{tran2019zero}
Ke~M Tran and Arianna Bisazza. 2019.
\newblock Zero-shot dependency parsing with pre-trained multilingual sentence
  representations.
\newblock In \emph{Proceedings of the 2nd Workshop on Deep Learning Approaches
  for Low-Resource NLP (DeepLo 2019)}, pages 281--288.

\bibitem[{Vilares et~al.(2016)Vilares, G{\'o}mez-Rodr{\'\i}guez, and
  Alonso}]{vilares2016one}
David Vilares, Carlos G{\'o}mez-Rodr{\'\i}guez, and Miguel~A Alonso. 2016.
\newblock One model, two languages: training bilingual parsers with harmonized
  treebanks.
\newblock In \emph{Proceedings of the 54th Annual Meeting of the Association
  for Computational Linguistics (Volume 2: Short Papers)}, pages 425--431.

\bibitem[{Wu and Dredze(2019)}]{wu2019beto}
Shijie Wu and Mark Dredze. 2019.
\newblock Beto, bentz, becas: The surprising cross-lingual effectiveness of
  bert.
\newblock In \emph{Proceedings of the 2019 Conference on Empirical Methods in
  Natural Language Processing and the 9th International Joint Conference on
  Natural Language Processing (EMNLP-IJCNLP)}, pages 833--844.

\bibitem[{Wu et~al.(2016)Wu, Schuster, Chen, Le, Norouzi, Macherey, Krikun,
  Cao, Gao, Macherey et~al.}]{wu2016google}
Yonghui Wu, Mike Schuster, Zhifeng Chen, Quoc~V Le, Mohammad Norouzi, Wolfgang
  Macherey, Maxim Krikun, Yuan Cao, Qin Gao, Klaus Macherey, et~al. 2016.
\newblock Google's neural machine translation system: Bridging the gap between
  human and machine translation.
\newblock \emph{arXiv preprint arXiv:1609.08144}.

\bibitem[{Zeman and Resnik(2008)}]{zeman2008cross}
Daniel Zeman and Philip Resnik. 2008.
\newblock Cross-language parser adaptation between related languages.
\newblock In \emph{Proceedings of the IJCNLP-08 Workshop on NLP for Less
  Privileged Languages}.

\bibitem[{Zhang and Barzilay(2015)}]{zhang2015hierarchical}
Yuan Zhang and Regina Barzilay. 2015.
\newblock Hierarchical low-rank tensors for multilingual transfer parsing.
\newblock In \emph{Proceedings of the 2015 Conference on Empirical Methods in
  Natural Language Processing}, pages 1857--1867.

\end{thebibliography}
\bibliographystyle{acl_natbib}

\newpage
\clearpage
\appendix

\section{Appendix}
\subsection{Experimental Details}
\label{app:imp}

\begin{table}[t]
\centering
\small
\begin{tabular}{@{}ll@{}}
\toprule
Hyper-Parameter & Value \\ \midrule
Dependency tag dimension   & 256      \\ 
Dependency arc dimension   & 768      \\
Optimizer                  & Adam     \\
$\beta_1,~\beta_2$         & 0.9, 0.99 \\
Weight decay               & 0.01     \\
Label smoothing            & 0.03    \\
Dropout                    & 0.5      \\
BERT dropout               & 0.2      \\
Mask probability           & 0.2      \\
Batch size                 & 32       \\
Epochs                     & 80       \\
Base learning rate         & $\textrm{1e}^{-3}$     \\
BERT learning rate         & $\textrm{5e}^{-5}$     \\
LR warm up ratio & $\textrm{1}/\textrm{80}$    \\
\midrule
Adapter size            & 256 \\
Language embedding size & 32 \\
\bottomrule
\end{tabular}
\caption{Hyper-parameter setting}
\label{tab:hyperparams}
\end{table}


\paragraph{Implementation} 
UDapter's implementation is based on UDify \cite{kondratyuk201975}. \au{We use the same hyper-parameters setting optimized in UDify without applying a new hyper-parameter search.} Together with the additional \textit{adapter size} and \textit{language embedding size} that are picked manually by parsing accuracy, hyper-parameters are given in Table \ref{tab:hyperparams}. Note that, \AB{to give a fair chance to the adapter-only baseline (see \S\ref{sec:exp})}, we used 1024 as adapter size unlike that of the final UDapter (256). For fair comparison, mono-udify and multi-udify are re-trained on the concatenation of 13 high-resource languages for only dependency parsing. Besides, we did not use a layer attention for both our model and the baselines.

\paragraph{Training Time and Model size} 
Comparing to UDify, UDapter has a similar training time. An epoch over the full training set takes approximately 27 and 30 minutes in UDify and UDapter respectively on a Tesla V100 GPU. In terms of number of \textit{trainable} parameters, 
\ready{UDify has 191M total number of parameters whereas UDapter uses 550M parameters in total, 302M for adapters (32x9.4M) and 248M for biaffine attention (32x7.8M), since the parameter generator network (CPG) multiplies the tensors with language embedding size (32). Note that for multilingual training, UDapter's parameter cost depends only on language embedding size regardless of number of languages, therefore it highly scalable with an increasing number of languages for larger experiments.}
%
Finally, monolingual UDify models are trained separately so the total number of parameters for 13 languages is 2.5B (13x191M).

\subsection{Zero-Shot Results}
\label{app:full-scores}

Table \ref{tab:full-lr-results} shows LAS scores on all 30 low-resouce languages for UDapter, original UDify \cite{kondratyuk201975}, and re-trained `multi-udify'. Languages with `*' are not included in mBERT training data. Note that original UDify is trained on all available UD treebanks from 75 languages. For the zero-shot languages, we obtained original UDify scores by running the pre-trained model.

\begin{table}[t]
\centering
\small
\begin{tabular}{@{}lccc|c@{}} 
\toprule
     & orig.udify & multi-udify & udapter & udap.-proxy\\ \midrule
aii* & 9.1            & 8.4         & \textbf{14.3}  & 8.2 (ar)  \\
akk* & 4.4            & 4.5         & \textbf{8.2}   & 9.1 (ar)   \\
am*  & 2.6            & 2.8         & \textbf{5.9}    & 1.1 (ar) \\
be   & \textbf{81.8}           & 80.1        & 79.3   & 69.9 (ru) \\
bho*$(\dagger)$ & 35.9           & 37.2        & \textbf{37.3}   & 35.9 (hi) \\
bm*  & 7.9            & \textbf{8.9}         & 8.1    & \scriptsize{3.1 \textsc{(ctr)}} \\
br*  & 39.0           & \textbf{60.5}        & 58.5   & \scriptsize{14.3 \textsc{(ctr)}} \\
bxr* & 26.7           & 26.1        & \textbf{28.9}   & \scriptsize{9.1 \textsc{(ctr)}} \\
cy   & 42.7           & 53.6        & \textbf{54.4}   & \scriptsize{9.8 \textsc{(ctr)}} \\
fo*  & 59.0           & 68.6        & \textbf{69.2}   & 64.1 (sv) \\
gsw* & 39.7           & 43.6        & \textbf{45.5}   & 23.7 (en) \\
gun*$(\dagger)$ & 6.0            & \textbf{8.5}         & 8.4    & \scriptsize{2.1 \textsc{(ctr)}} \\
hsb* & \textbf{62.7}           & 53.2        & 54.2   & 44.4 (ru) \\
kk   & \textbf{63.6}           & 61.9        & 60.7   & 45.1 (tr) \\
kmr*$(\dagger)$ & \textbf{20.2}           & 11.2        & 12.1   & \scriptsize{4.7 \textsc{(ctr)}} \\
koi* & 22.6           & 20.8        & \textbf{23.1}   & \scriptsize{6.5 \textsc{(ctr)}} \\
kpv*$(\dagger)$ & \textbf{12.9}           & 12.4        & 12.5   & \scriptsize{4.7 \textsc{(ctr)}} \\
krl* & 41.7           & \textbf{49.2}        & 48.4   & 45.6 (fi) \\
mdf* & 19.4           & 24.7        & \textbf{26.6}   & \scriptsize{8.7 \textsc{(ctr)}} \\
mr   & \textbf{67.0}           & 46.4        & 44.4   & 29.6 (hi) \\
myv*$(\dagger)$ & 16.6           & 19.1        & \textbf{19.2}   & \scriptsize{6.3 \textsc{(ctr)}} \\
olo* & 33.9           & 42.1        & \textbf{43.3}   & 41.1 (fi) \\
pcm*$(\dagger)$ & 31.5           & 36.1        & \textbf{36.7}   & \scriptsize{5.6 \textsc{(ctr)}} \\
sa*  & 19.4           & 19.4        & \textbf{22.2}   & 15.1 (hi) \\
ta~$(\dagger)$   & \textbf{71.4}           & 46.0        & 46.1   & \scriptsize{12.3 \textsc{(ctr)}} \\
te~$(\dagger)$   & \textbf{83.4}           & 71.2        & 71.1   & \scriptsize{23.1 \textsc{(ctr)}} \\
tl   & 41.4           & 62.7        & \textbf{69.5}   & \scriptsize{14.1 \textsc{(ctr)}} \\
wbp* & 6.7            & 9.6        & \textbf{12.1}    & \scriptsize{4.8 \textsc{(ctr)}}  \\
yo   & 22.0           & 41.2        & \textbf{42.7} & \scriptsize{10.5 \textsc{(ctr)}}  \\
yue* & 31.0           & 30.5        & \textbf{32.8}  & 24.5 (zh)  \\ \midrule
avg  & 34.1           & 35.3        & \textbf{36.5}  & 20.4 \\ \bottomrule
\end{tabular}
\caption{LAS results of UDapter and UDify models \cite{kondratyuk201975} for all low-resource languages.~`*' shows languages not present in mBERT training data.~Additionally, \ready{$(\dagger)$ indicates languages where no significant difference between UDapter and \textit{multi-udify} by significance testing.}~For udapter-proxy, chosen proxy language is given between brackets. \textsc{ctr} means centroid language embedding.}
\label{tab:full-lr-results}
\end{table}

\subsection{Language Details}
\label{app:langs}
Details of training and zero-shot languages such as language code, data size (number of sentences), and family are given in Table \ref{tab:hr-lang-app} and Table \ref{tab:lr-langs-app}. 

\begin{table*}[h]
\centering\small
\begin{tabular}{@{}lllllll@{}}
\toprule
Language & Code & Treebank  & Family               & Word Order & Train & Test\\ \midrule
Arabic   & ar   & PADT      & Afro-Asiatic, Semitic & VSO        & 6.1k  & 680 \\
Basque   & eu   & BDT       & Basque               & SOV        & 5.4k   & 1799\\
Chinese  & zh   & GSD       & Sino-Tibetan         & SVO        & 4.0k  & 500 \\
English  & en   & EWT       & IE, Germanic          & SVO        & 12.5k & 2077 \\
Finnish  & fi   & TDT       & Uralic, Finnic        & SVO        & 12.2k & 1555 \\
Hebrew   & he   & HTB       & Afro-Asiatic, Semitic & SVO        & 5.2k  & 491 \\
Hindi    & hi   & HDTB      & IE, Indic             & SOV        & 13.3k & 1684 \\
Italian  & it   & ISDT      & IE, Romance           & SVO        & 13.1k & 482 \\
Japanese & ja   & GSD       & Japanese             & SOV        & 7.1k  & 551 \\
Korean   & ko   & GSD       & Korean               & SOV        & 4.4k  & 989 \\
Russian  & ru   & SynTagRus & IE, Slavic            & SVO        & 15k* & 6491 \\
Swedish  & sv   & Talbanken & IE, Germanic          & SVO        & 4.3k & 1219 \\
Turkish  & tr   & IMST      & Turkic, Southwestern  & SOV        & 3.7k & 975 \\ \bottomrule
\end{tabular}
\caption{Training languages that are from UD 2.3 \cite{11234/1-2895} with the details including treebank name, family, word order and data size of training and test sets.}
\label{tab:hr-lang-app}
\end{table*}

\begin{table*}[h]
\centering\small
\begin{tabular}{@{}lllll@{}}
\toprule
Language      & Code & Treebank(s)      & Family        & Test                  \\ \midrule
Akkadian      & akk  & PISANDUB      & Afro-Asiatic, Semitic    & 1074       \\
Amharic       & am   & ATT           & Afro-Asiatic, Semitic    & 101       \\
Assyrian      & aii  & AS            & Afro-Asiatic, Semitic    & 57       \\
Bambara       & bm   & CRB           & Mande                    & 1026       \\
Belarusian    & be   & HSE           & IE, Slavic               & 253        \\
Bhojpuri      & bho  & BHTB          & IE, Indic                & 254        \\
Breton        & br   & KEB           & IE, Celtic               & 888       \\
Buryat        & bxr  & BDT           & Mongolic                 & 908       \\
Cantonese     & yue  & HK            & Sino-Tibetan             & 1004       \\
Erzya         & myv  & JR            & Uralic, Mordvin          & 1550       \\
Faroese       & fo   & OFT           & IE, Germanic             & 1207       \\
Karelian      & krl  & KKPP          & Uralic, Finnic           & 228        \\
Kazakh        & kk   & KTB           & Turkic, Northwestern     & 1047        \\
Komi Permyak  & koi  & UH            & Uralic, Permic           & 49       \\
Komi Zyrian   & kpv  & LATTICE, IKDP  & Uralic, Permic          & 210         \\
Kurmanji      & kmr  & MG            & IE, Iranian              & 734        \\
Livvi         & olo  & KKPP          & Uralic, Finnic           & 106        \\
Marathi       & mr   & UFAL          & IE, Indic                & 47        \\
Mbya Guarani  & gun  & THOMAS, DOOLEY & Tupian                  & 98        \\
Moksha        & mdf  & JR            & Uralic, Mordvin          & 21        \\
Naija         & pcm  & NSC           & Creole                   & 948       \\
Sanskrit      & sa   & UFAL          & IE, Indic                & 230        \\
Swiss G.      & gsw  & UZH           & IE, Germanic             & 100        \\
Tagalog       & tl   & TRG           & Austronesian, Central Philippine & 55 \\
Tamil         & ta   & TTB           & Dravidian, Southern      & 120        \\
Telugu        & te   & MTG           & Dravidian, South Central & 146        \\
Upper Sorbian & hsb  & UFAL          & IE, Slavic               & 623        \\
Warlpiri      & wbp  & UFAL          & Pama-Nyungan             & 54       \\
Welsh         & cy   & CCG           & IE, Celtic               & 956        \\
Yoruba        & yo   & YTB           & Niger-Congo, Defoid      & 100        \\ \bottomrule
\end{tabular}
\caption{Zero-shot languages are selected from UD 2.5 to increase the number of languages in the experiments. Language details include treebank name, family and test size for zero-shot experiments.}
\label{tab:lr-langs-app}
\end{table*}

\end{document}